\documentclass[runningheads]{llncs}
\usepackage{footmisc}
\usepackage{graphicx}
\usepackage{hyperref}
\begin{document}
\title{Avoiding Echo-Responses in a Retrieval-Based Conversation System}
%
%
\author{Denis Fedorenko\orcidID{0000-0002-4677-3872} \and
Nikita Smetanin\orcidID{0000-0002-5990-3840} \and
Artem Rodichev\orcidID{0000-0002-9448-2236}}

\authorrunning{D. Fedorenko et al.}

\institute{Replika.ai @ Luka, Inc.\\
\email{\{denis,nikita,artem\}@replika.ai}}

\maketitle              
\begin{abstract}
Retrieval-based conversation systems generally tend to highly rank responses that are semantically similar or even identical to the given conversation context. While the system's goal is to find the most appropriate response, rather than the most semantically similar one, this tendency results in low-quality responses. We refer to this challenge as the echoing problem. To mitigate this problem, we utilize a hard negative mining approach at the training stage. The evaluation shows that the resulting model reduces echoing and achieves better results in terms of Average Precision and Recall@N metrics, compared to the models trained without the proposed approach.

\keywords{Dialog modeling \and Response selection \and Lexical repetition \and Hard negative mining \and End-to-end learning}
\end{abstract}
\section{Introduction}

The task of a retrieval-based conversation system is to select the most appropriate response from a set of responses given the input context of a conversation. The context is typically an utterance or a sequence of utterances produced by a human or by the system itself.
Most of the state-of-the-art approaches to retrieval-based conversation systems are based on deep neural networks (NNs) \cite{zhou2016multi,DBLP:journals/corr/WuWZL16}. Under these approaches, the typical response selection pipeline consists of the following steps \cite{Chen:2017:SDS:3166054.3166058}:

\begin{enumerate}
\item Encode the given context and pre-defined response candidates into numeric vectors, or thought vectors, using NNs;
\item Compute the value of a matching function (matching score) for each pair consisting of a context vector and each response candidate;
\item Select the response candidate with the highest matching score.
\end{enumerate}

During step 1, in order to obtain thought vectors that fairly represent semantics of input contexts and responses, the conversation model is preliminarily trained to return high matching scores for true context-response pairs and low for false ones.

The challenge we faced while building the above pipeline was that the resulting model often returned high matching scores for semantically similar contexts and responses. Consequently, the model frequently repeated or rephrased input contexts instead of giving quality responses.

Consider the following conversations:
\renewcommand{\labelenumi}{\Alph{enumi}.}
\begin{enumerate}
\item \label{e1} Context:~``What is the purpose of living?'' \\ Response:~``What is the purpose of existence?''
\item \label{e2} Context:~``What is the purpose of living?'' \\ Response:~``It's a very philosophical question.''
\end{enumerate}

The effect of rephrasing, or echoing, in conversation A in contrast to the appropriate response in conversation B can be explained by the above pipeline. It is a result of the fact that contexts and responses often contain the same concepts \cite{wang2013dataset,jurafsky2017dialog}, hence during training on conversational datasets the NNs simply end up trying to fit the semantics of the input. The similar effect, named ``lexical repetition'', was also observed in \cite{ritter2011data}.

In this paper, we suggest a simple and natural solution to the echoing problem for end-to-end retrieval-based conversation systems. Our solution is based on a widely used hard negative mining approach \cite{DBLP:journals/corr/SchroffKP15}, which forces the conversation model to produce low matching scores for similar contexts and responses.

The paper is organized as follows. First, we describe the hard negative mining method and how we utilize it to overcome the echoing problem. Then, we introduce the evaluation metrics, our results and benchmarks for the echoing problem. We also provide the evaluation dataset used in the experiments for further research.

\section{Hard Negative Mining}

\begin{figure}[b!]
\centerline{\includegraphics[width=0.70\textwidth]{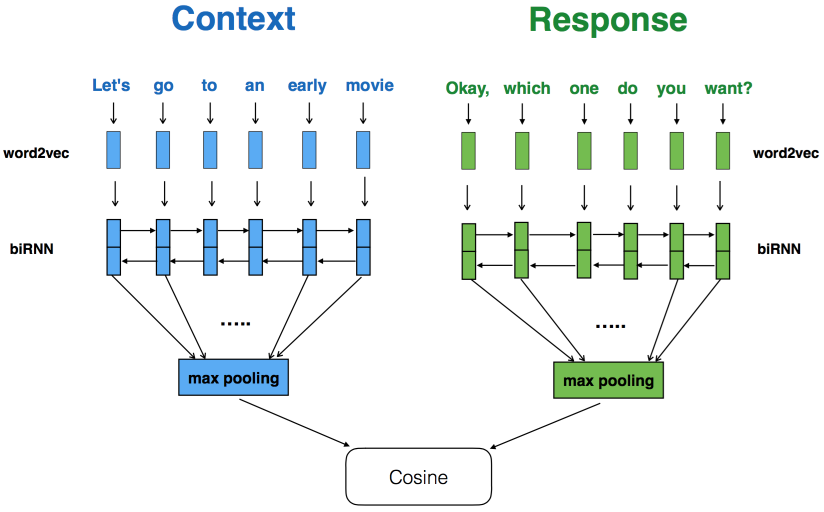}}
\caption{Conversation model architecture used in the experiments (see Section~\ref{sec:experiments}). It has two bidirectional LSTMs that encode a context and a response independently. Input words are represented as word2vec embeddings. The output matching function is a cosine similarity.
}
\label{fig:dialog_model}
\end{figure}

Let $D=\{(c_i, r_i)\}$, $i \in \{1..N\}$ be a dataset of conversational context-response pairs, where $c_i$, $r_i$ -- $i$-th context and response, respectively.

Our goal is to build a conversation model $M: \mathit{(context, response)} \rightarrow \rm I\!R$
that satisfies the following condition:
\begin{equation}
M(c_i, r_i) > M(c_i, r_j)
\end{equation}
$\forall i, j \neq i$ and $r_j$ is not an appropriate response for $c_i$. In other words, the resulting model should return a higher matching score for appropriate responses than for inappropriate ones.

To train this model, we also need false context-response pairs as negative examples in addition to the positive ones presented in $D$. Consider two approaches to obtain the negative pairs: random sampling and hard negative mining. Under the first approach, we randomly select $r_j$ from $D$ for each $c_i$. If $D$ is large and diverse enough, then a randomly selected $r_j$ is almost always inappropriate for a corresponding $c_i$.

In contrast to random sampling, hard negative mining imposes a special constraint on responses selected as negatives. Let $M_0$ be a conversation model trained on random pairs used as negative training examples. Then, we search for a new set of negative pairs $(c_i, r_j)$, so that their matching score satisfies the following condition:
\begin{equation}
M_0(c_i, r_i)-M_0(c_i, r_j) \leq m
\end{equation}
where $m$ is a margin (hyperparameter) between the scores of positive and negative pairs \cite{feng2015applying}. The new set of pairs is used to train the next model $M_1$, which, in turn, used to search for negative pairs to train $M_2$, and so on \cite{canevet2014efficient}.

The intuitive idea behind hard negative mining is to select only negatives that have relatively high matching scores, and thus can be interpreted as errors of the conversation model. As a result, the model converges faster compared to random sampling \cite{DBLP:journals/corr/SchroffKP15}.

Following this intuition, we can solve the echoing problem by considering contexts as possible responses, therefore the pairs $(c_i, c_i)$ can be selected as hard negatives. In the next section, we demonstrate that this approach can ultimately prevent the conversation model from assigning a high rank to responses that are similar to contexts.

\section{Experiments}
\label{sec:experiments}

For our experiments, we implement a model similar to Basic QA-LSTM described in \cite{DBLP:journals/corr/TanXZ15} (see Figure~\ref{fig:dialog_model}). It has two bidirectional LSTMs of size 2048 (1024 units in each direction), with separate sets of weights that encode a context and a response independently. We use a max pooling operation to calculate final thought vectors of these LSTMs. We use a cosine similarity as the output matching function. We represent input words as embeddings of size 256, which are initialized by the pre-trained word2vec vectors \cite{DBLP:journals/corr/abs-1301-3781} and are not updated further during the model training. Word sequences longer than 20 words are trimmed from the right, and the context encoder is fed with only one dialog step at a time.

\subsection{Models}

In order to study the impact of hard negative mining on the echoing problem, we train three models using the following strategies: random negative sampling ($\mathit{RN}$), hard negative mining based on responses only ($\mathit{HN_r}$), and hard negative mining based on both responses and contexts ($\mathit{HN_{r+c}}$). We also consider the following baseline approach ($\mathit{BL}$): we use $\mathit{RN}$ model to rank responses in the testing stage and then just filter out responses equal to the given context.

\subsection{Datasets}
\label{sec:datasets}

We train the models on 79M of tweet-reply pairs from a Twitter data archive \footnote{\href{https://archive.org/details/twitterstream}{https://archive.org/details/twitterstream}}. 

We perform an evaluation based on our own dataset \footnote{\href{https://github.com/lukalabs/replika-research/tree/master/context-free-dataset}{https://github.com/lukalabs/replika-research/tree/master/context-free-dataset}\label{fn:testset}}. This dataset consists of 759 context-response pairs from human text conversations, where context and response both consist of a single sentence (see Table~\ref{tab:dataset_sample}). We split the dataset into validation and test subsets consisting of 250 and 509 pairs, respectively. We use this dataset because it is clear, diverse and covers multiple topics of real-life conversations. Also we find it suitable for validating the echoing problem, as well as for estimating the overall model quality.

\begin{table*}[t]
\centering
\small
\begin{tabular}{ |c|c| }
 \hline
 \textbf{context} & \textbf{response} \\ 
 \hline
 What happened to your car? & I got a dent in the parking lot. \\ 
 The Beatles are the best. & They are the best musical group ever. \\ 
 Do you want to go fishing? & Yes. That's a good idea. \\
 What do you think about Britney Spears? & Oh, she's a great singer. \\
 White coffee, no sugar please. & Here you are. \\
 I'm joining the army. & You're kidding. You might get killed. \\
 \hline
\end{tabular}
\caption{\label{tab:dataset_sample} Evaluation dataset sample (see Section~\ref{sec:datasets})}
\end{table*}

\subsection{Training}

The models are trained with the Adam optimizer \cite{kingma2014adam} with the size of mini-batches set to 512.
Intermediate models that show the highest values of the Average Precision metric on the validation set (see Section~\ref{sec:metrics}) are selected as the resulting models.

We use a triplet loss \cite{feng2015applying} as an objective function:
\begin{equation}
max(0, m - M(c_i, r_i) + M(c_i, r_j))
\end{equation}
where the margin $m$ is set to 0.05. For each positive pair $(c_i, r_i)$, a negative $(c_i, r_j)$ is only selected within the current mini-batch using an intermediate model $M$ trained by the moment of this batch. We only select the hard negative $r_j$ with the highest matching score $M(c_i, r_j)$ satisfying the following condition:
\begin{equation}
0 \leq M(c_i, r_i) - M(c_i, r_j) \leq m
\end{equation}
The constraint $0 \leq M(c_i, r_i) - M(c_i, r_j)$ is used to filter out the ``hardest'' negatives, which in practice affect convergence and lead to bad local optima \cite{DBLP:journals/corr/SchroffKP15}. 

We noticed that while training the $\mathit{HN_{r+c}}$ model, the fraction of $(c_i, c_i)$ negative pairs constitute up to 50\% of the mini-batch.

\subsection{Evaluation Methodology and Metrics}
\label{sec:metrics}

\begin{table}[b]
\centering
\small
\begin{tabular}{ |c|c|c|c|c| }
   \hline
    & $\mathit{RN}$ & $\mathit{BL}$ & $\mathit{HN_r}$ & $\mathit{HN_{r+c}}$ \\ 
   \hline
   Average Precision & 0.12 & 0.16 & 0.13 & \textbf{0.17} \\ 
   Recall@2 & 0.18 & 0.26 & 0.23 & \textbf{0.29} \\
   Recall@5 & 0.36 & 0.37 & 0.4 & \textbf{0.43} \\ 
   Recall@10 & 0.45 & 0.48 & \textbf{0.54} & 0.53 \\ 
   $\mathit{rank_{context}}$ & 0.9 & - & 0.49 & \textbf{19.43} \\ 
   $\mathit{diff_{top}}$ & 0.008 & - & 0.01 & \textbf{0.07} \\ 
   $\mathit{diff_{response}}$ & -0.15 & - & -0.25 & \textbf{-0.09} \\ 
 \hline
\end{tabular}
\caption{\label{tab:quality_metrics} Evaluation results based on the context-response test set~\footref{fn:testset}.
$\mathit{RN}$ -- random negatives model; $\mathit{BL}$ -- $\mathit{RN}$ model with responses filter; $\mathit{HN_r}$ -- hard negatives model based on responses only; $\mathit{HN_{r+c}}$ -- proposed hard negatives model based on both responses and contexts. The metrics are described in Section~\ref{sec:metrics}
}
\end{table}

For each $\mathit{context_i}$ from the evaluation set, we compute matching scores for all available pairs $\mathit{(context_i, answer)}$, where $\mathit{answer}$ comes not only from the responses, but also from the all available contexts. To evaluate these results, we sort the answers by the matching score in descending order and compute the following metrics: Average~Precision \cite{Manning:2008:IIR:1394399}, Recall@2, Recall@5, and Recall@10 \cite{DBLP:journals/corr/LowePSP15}. The last three metrics are indicator functions that return 1, if the ground-truth response occurs in the top 2, 5 and 10 candidates, respectively. We also introduce the context echoing metrics:

\begin{itemize}
\item[\textbullet] $\mathit{rank_{context}}$ -- position (starting from zero) of the input context in the sorted results. The greater the rank, the less the model tends to return the input context among the top results
\item[\textbullet] $\mathit{diff_{top}}$ -- difference between the top result score and the input context score. The greater the difference, the less the model tends to return relatively high scores for the context
\item[\textbullet] $\mathit{diff_{response}}$ -- difference between the ground-truth response score and the input context score. The greater the difference, the less the model tends to return similar scores for the ground-truth response and for the context
\end{itemize}

For each metric, we compute the overall quality as an average across all test contexts. Note that for $BL$ model we don't present context echoing metrics, since echo-responses are filtered out from the results in this approach.

\subsection{Results}

\begin{table*}[t!]
\centering
\small
\begin{tabular}{ |p{4.0cm}|p{4.0cm}|p{4.0cm}| }
   \hline
   \multicolumn{1}{|c|}{$\mathit{RN}$} & \multicolumn{1}{|c|}{$\mathit{HN_r}$} & \multicolumn{1}{|c|}{$\mathit{HN_{r+c}}$} \\ 
   \hline
   
   \multicolumn{3}{|l|}{\textbf{Input:} What is the purpose of dying?} \\
   \hline
   1. What is the purpose of dying? & 1. What is the purpose of dying? & 1. To have a life. \\
   2. The victim hit his head on the concrete steps and died. & 2. What is the purpose of living? & 2.~When you die and go to heaven, they will offer you beer or cigarettes. \\
   3. To have a life. & 3. What is the purpose of existence? & 3. It is to find the answer to the question of life. \\
   \hline
   \hline
   
   \multicolumn{3}{|l|}{\textbf{Input:} What are your strengths?} \\
   \hline
   1. What are your strengths? & 1. What are your strengths? & 1. Lust, greed, and corruption. \\
   2. Lust, greed, and corruption. & 2.~What are your three weaknesses? & 2. I'm a robot. a machine. 100\% ai. no humans involved \\
   3. A star. & 3. What do you think about creativity? & 3. Dunno. i mean, i'm a robot, right? robots don't have a gender usually \\
   \hline
   \hline
   
   \multicolumn{3}{|l|}{\textbf{Input:} I can't wait until i graduate.} \\
   \hline
   1. I can't wait until i graduate. & 1. I can't wait until i graduate. & 1. What college do you go to? \\
   2. What college do you go to? & 2. What college do you go to? & 2. School is hard this year. \\
   3. School is hard this year. & 3. How many jobs have you had since leaving university? & 3. What subjects are you taking? \\
   \hline
  \hline
   
  \multicolumn{3}{|l|}{\textbf{Input:} Lunch was delicious.} \\
  \hline
  1. Lunch was delicious. & 1. Lunch was delicious. & 1. Who did you go out with? \\
  2. I want to buy lunch. & 2. I want to buy lunch. & 2. So was i. \\
  3. Take me to dinner. & 3. This hot bread is delicious. & 3. What did you do today? \\
  \hline
  \hline
   
 \multicolumn{3}{|l|}{\textbf{Input:} You're crazy} \\
 \hline
 1. You're crazy & 1. You're crazy & 1. Am i? \\
 2. Am i? & 2. Am i? & 2. You're crazy \\
 3. I sure am. & 3. Why? what have i done? & 3. I sure am. \\
 \hline
\end{tabular}
\caption{\label{tab:model_output} Top 3 responses for a few input contexts sorted by matching score.
$\mathit{RN}$ -- random negatives model; $\mathit{HN_r}$ -- hard negatives model based on responses only; $\mathit{HN_{r+c}}$ -- proposed hard negatives model based on both responses and contexts. The test set is described in Section~\ref{sec:datasets}
}
\end{table*}

The results of the evaluation based on the test set are presented in Table~\ref{tab:quality_metrics}. As we can see, the proposed $\mathit{HN_{r+c}}$ model achieves the highest values in almost all metrics compared to other approaches. According to $\mathit{rank_{context}}$, it turns out that this model does not tend to highly rank input contexts and have them in the top response candidates. Still, according to the $\mathit{diff_{response}}$ metric, the average score of a ground-truth response is lower than the score of a context, which means that the context can be ranked higher than the ground-truth response.

We also studied the model's output. Examples of top-ranked responses for different contexts are presented in Table~\ref{tab:model_output}. As we can see, oftentimes the $\mathit{RN}$ and $\mathit{HN_r}$ models select identical or very similar responses, while the proposed $\mathit{HN_{r+c}}$ model selects appropriate responses that are not necessarily semantically similar to the context. Based on this observation, we suggest that the proposed model filters out not only exact copies of the context, but also candidates with similar semantics.
Moreover, in some cases the model selects semantically similar responses which are, at the same time, appropriate for a given context. See Table~\ref{tab:hello_output} with the top results for the context ``Hello."

\begin{table}[b]
\centering
\small
\begin{tabular}{ |p{2cm}|p{2.5cm}| }
  \hline
  \textbf{matching score} & \textbf{response} \\
  \hline
  0.45 & Hey, sweetie \\
  0.44 & How's life ? \\
  0.43 & Hello \\
  \hline
\end{tabular}
\caption{\label{tab:hello_output} Top responses of the $\mathit{HN_{r+c}}$ model for the context ``Hello."}
\end{table}

\section{Related Work}

In the previous works on dialog systems there was not enough attention paid to the echoing problem. The possible reason for this are ``soft'' evaluation conditions: test samples are constructed from a relatively small number of negative responses \cite{DBLP:journals/corr/LowePSP15,feng2015applying,DBLP:journals/corr/WuWZL16} which usually do not ``echo'' the test context. In \cite{ritter2011data} the ``lexical repetition'' is regularized by utilizing a word overlap feature during training a SMT-based dialog system. In \cite{wang2013dataset,DBLP:journals/corr/WuWZL16,serban2017deep} the echoing is avoided by considering only responses the dataset's contexts of which have high TF-IDF similarity with the given context. However, the latter approach is not applicable if only a set of responses is available for ranking during the testing stage, which can be the case for some domains and applications \cite{yan2016docchat}.

\section{Conclusion}

In this study, we applied a hard negative mining approach to train a retrieval-based conversation system to find a solution to the echoing problem, that is, to reduce inappropriate responses that are identical or too similar to the input context. In addition to responses, we consider contexts themselves as possible hard negative candidates. The evaluation shows that the resulting model avoids echoing the input context, tends to select candidates that are more appropriate as responses and achieves better results in terms of Average Precision and Recall@N metrics compared to the models trained without the proposed approach.

\section{Acknowledgments}

This is a pre-print of an article published in ``Artificial Intelligence and Natural Language'' (7th International Conference, AINL 2018, St. Petersburg, Russia, October 17-19, 2018). The final authenticated version is available online at: \href{https://doi.org/10.1007/978-3-030-01204-5_9}{https://doi.org/10.1007/978-3-030-01204-5\_9}

\bibliography{bibliography}
\bibliographystyle{splncs04}

\end{document}